# A Facial Feature Discovery Framework for Race Classification Using Deep Learning

Khalil Khan[1], Jehad Ali[2], Irfan Uddin[3], Sahib Khan[4], and Byeong-hee Roh[2, *]

[1]Department of Information Technology and Computer Science, Pak-Austria Institute of Applied Science and Technology, Haripur, Pakistan
[2]Dept. of Computer Engineering and Dept. of AI Convergence Network, Suwon, South Korea
[3]Departmet of Computer Science, Superior College, Lahore, Pakistan
[4]Department of Electronics and Telecommunications, Politecnico di Torino, Torino, Italy
[*]Corresponding Author: Byeong-hee Roh. Email: bhroh@ajou.ac.kr

**Abstract:** Race classification is a long-standing challenge in the field of face image analysis. The investigation of salient facial features is an important task to avoid processing all face parts. Face segmentation strongly benefits several face analysis tasks, including ethnicity and race classification. We propose a race-classification algorithm using a prior face segmentation framework. A deep convolutional neural network (DCNN) was used to construct a face segmentation model. For training the DCNN, we label face images according to seven different classes, that is, nose, skin, hair, eyes, brows, back, and mouth. The DCNN model developed in the first phase was used to create segmentation results. The probabilistic classification method is used, and probability maps (PMs) are created for each semantic class. We investigated five salient facial features from among seven that help in race classification. Features are extracted from the PMs of five classes, and a new model is trained based on the DCNN. We assessed the performance of the proposed race classification method on four standard face datasets, reporting superior results compared with previous studies.

**Keywords:** Deep learning; facial feature; face analysis; learning race; race classification

## 1 Introduction

Race classification is a difficult task in computer vision (CV). The analysis of race based on facial features is a popular topic in face recognition and CV communities [1]–[3]. With increasing globalization, face recognition has many applications in customs checks, border control, public security, and many other fields. Moreover, it is also a research branch in the field of physical anthropology. Facial features are greatly influenced by the environment, genes, society, and other factors. The genes of one race group are hardly distinguishable from other groups due to various gene fragments. Hence, facial features among different races are similar [4]. In our proposed research work, different facial features are analyzed for race classification, and based on data mining, some important facial cues that help in race classification are suggested. Our work may also be helpful in the field of anthropology, which investigates facial feature evolution, as in [5].

In the 19th century, it was proposed that face analysis tasks have a close relationship with each other. Specifically, gaze estimation and head pose estimation were considered to be closely related to each other [6]. Similarly, in [7], it was proposed that various face analysis tasks are related to each other. In the initial phase of these tasks, the face is segmented into different dense semantic classes; then, face segmentation information is used for head pose prediction. Our work is also inspired by this [6], [7]. However, we believe that comparatively better results can be attained if complete and accurate face segmentation is performed and subsequently used for various hidden facial cues. Our proposed work represents a continuation of the long-term research of [8], [9]. A given face image is segmented into seven classes: mouth, skin, nose, skin, hair, brow, back, and eyes. We first developed a face segmentation framework by manually labeling the face images. Recently, there has been a shift in state-of-the-art methods from conventional machine learning to novel deep learning methods. We also developed a face segmentation model using a deep convolutional neural network (DCNN).

Race classification has already been addressed by CV researchers in different ways [1]–[3], [10], [11]. However, most of these methods either use landmark localization [10]–[12] or consider a face image as a one-dimensional feature vector [2], [3]. Between these two methods, landmarks localization shows comparatively better performance. However, landmark localization is in itself a daunting challenge in some critical scenarios. For example, landmark extraction methods completely fail if the facial expression is complicated, some changes in face rotation or lighting conditions occur, and if the image is taken from a distance. In all these and many other cases, landmark localization completely fails. Unlike these two methods, we approach race classification in a completely different way that does not depend on facial landmarks or consider faces as one-dimensional vectors.

We believe that the performance of race classification can be improved if information about various face parts in the form of segmented images is provided. Psychology literature confirmed the same fact in [13], [14]. In summary, the task of race classification can be addressed in a better way if facial part information is extracted, and then, the extracted information is given to the race classification framework. Thus, unlike previous methods, our proposed method addresses race classification in a completely different way. We first segmented a face image into seven classes. The face segmentation model is built by utilizing the extracted features from DCNNs. We build a SoftMax classifier and use a probabilistic classification method to create PMs for each dense class. We use five out of seven features and extract information from the PMs to build a new DCNN-based classifier. To summarize, the significant contributions of our paper are as follows:

- We propose a face segmentation algorithm that segments a given face image into seven face classes.
- We develop a new race classification algorithm. The race classification method is based on information conveyed by a previously built face segmentation model.
- We conduct experiments on race classification using SOA datasets. We obtain better results than previous studies.

The remainder of this paper is structured as follows. In Section II, related work on face segmentation and race classification is described. The proposed face-parsing and race classification algorithms are discussed in Section III. We discuss the obtained results in Section IV and compare it with SOA. A summary of the paper with some promising future directions is presented in Section V.

## 2 Literature Review

We broadly classify face segmentation literature based on the methods used: local and global face parsing methods. Both these methods are discussed in the following paragraphs. In the last part of this section, we discuss studies related to race classification methods. Local based methods adapt the strategy of coarse to fine. Both global consistency and local precision are considered by these methods. Individual models are designed for different face components using these methods. A method presented by Luo et al. [15] trains a model that parses each face part, such as mouth, nose, skin, individually. In the same way, Zhou et al. [16] interlinked different CNNs by first localizing face parts. The method presented in [16] passes information in both coarse and fine directions. However, owing to the bi-directional level exchange of information, this method is computationally heavy. Better results were achieved in the work presented in [17], which combines RNNs and DCNNs in two successive stages.

The second class of methods known as global-based methods treat various facial part information globally. Compared with local-based methods, comparatively poor results are reported for these methods, as individual face part information is not targeted. In some research a specific spatial relationship between various face parts was investigated, for example, [18] and [19]. The work presented in [18] encodes the underlying layout information of the face. The authors present the idea of integrating conditional random fields (CRFs) with CNNs. Similarly, Jackson et al. [20] combined CNNs with boundary cues to confine various face regions. Facial landmarks are exploited as an initial step for face part localization. The proposed method in [20] also employs fully convolutional neural networks (FCNNs), claiming improved performance. Wei et al. [21] presented the idea of regulating receptive fields in a face-segmentation network.

A substantial literature exists on race classification; however, here, we will try to present maximum information about recently introduced race classification methods. Holistic race classification algorithms consider an image to be a one-dimensional feature vector, extracting important information from it. For example, [22]–[24] used neural networks and decision trees for race classification. In [25], the FERET dataset were used for experiments. Similarly, another method proposed in [26] used discriminant analysis. This method also classifies face images into two classes: Asian and non-Asian. Different kinds of classifiers have been tested to address race classification; for instance, [27] used a support vector machine (SVM) with the FERET dataset. An algorithm proposed in [28] combines logistic regression and binary pattern information (BPI) in a single framework called SPARK. In the proposed model, BPI is used for extracting features and Spark is used for classification. In [29], FERET and CAS-PEAL were used for the experimental work. However, the proposed work classifies face images into only two races, namely Asian and non-Asian. Manesh et al. [30] extracted features from face images through Gabor filtering and then performed classification with SVM. An interesting method proposed in [31] addresses race classification using skin information and features extracted from the lip area, using the Yale [32] and FERET datasets. The sole work that performed classification for five classes, including American, Asian, Caucasian, African, and European, is reported in [33]. Features from the face are extracted through wavelet transform and local binary patterns. K-nearest neighbors (KNN) are used as classification tools in the proposed work.

All the aforementioned methods were evaluated on small datasets. Xie et al. [34] performed experiments on the comparatively larger dataset MORPH2. Face information is extracted from the color features. The dataset contains three classes of images. Han et al. [35] proposed a method that uses biologically inspired information and a set of hierarchical classifiers. Two larger datasets, MORPH2 and PSCO, are used for experimentation.

Deep learning improves SOA results in many CV applications. The same is true for race classification. A method based on real time conditions is proposed in [36]. The authors of the work used DCNNs, claiming better results, and a computationally much better framework. Zhang et al. [37] proposed a method that uses stacked auto-encoding information for feature extraction along with a SoftMax classifier. Similarly, an extremely flexible method for race classification was proposed by Wei et al. [21]. Hypothesis CNN pooling is a method proposed in [38]. It uses the several object segment hypotheses; DCNN were used for feature extraction and SVM for classification.

## 3 Face Features Based Algorithm

### 3.1 DCNNS Parameters

The performance of the DCNN model is greatly affected by several factors. The size of the kernel used for CNNs and the pooling layer is significant. Similarly, the number of layers used and the filters in each layer must also be investigated. In our DCNNs, we use four convolutional layers (CL1–CL4), followed by pooling layers (PL1–PL4). For the pooling layers, we choose the maximum pooling method. The last portion contains two fully connected layers. For the activation function, we use the rectified linear unit (ReLU). Each convolutional layer is followed by a maximum pooling layer. We fix the down-sampling stride, size of the kernel, and feature map, as shown in Table 1. Some additional parameters of the DCNNs are listed in Table 2.

**Table 1:** Various parameters of the DCNN.

| Layer | Stride | Size of kernel | Feature maps | Output size |
|---|---|---|---|---|
| *Input face image* | - | - | - | 250×250 |
| *CL-1* | 2 | 5×5 | 96 | 124×124 |
| *PL-1* | 2 | 3×3 | 96 | 62×62 |
| *CL-2* | 2 | 5×5 | 256 | 30×30 |
| *PL-2* | 2 | 3×3 | 256 | 15×15 |
| *CL-3* | 1 | 5×5 | 316 | 12×12 |
| *PL-3* | 2 | 3×3 | 316 | 6×6 |
| *CL-4* | 2 | 5×5 | 512 | 4×4 |
| *PL-4* | 2 | 3×3 | 512 | 2×2 |

**Table 2:** DCNNs training parameters and values.

| DCNNs parameters | Values |
|---|---|
| Epochs | 50 |
| Learning rate | $10^{-5}$ |
| Momentum | 0.8 |
| Size of batch | 250 |

### *3.2 Face Segmentation*

We build a face segmentation module using DCNNs for each dataset. If the dataset does not provide cropped face images, we perform face detection with [39]. After face detection, we resize each image to 250×250. Our DCNN-based face parsing module and its detailed architecture are shown in Table 1. The proposed DCNN face segmentation model has three parts: convolutional layers, pooling layers, and the last two fully connected layers. The fully connected layers at the end perform the classification task. We trained one classifier for each dataset. For more details about deep learning methods and their different parameters, please refer to [40]. We build a face segmentation module for all four datasets individually.

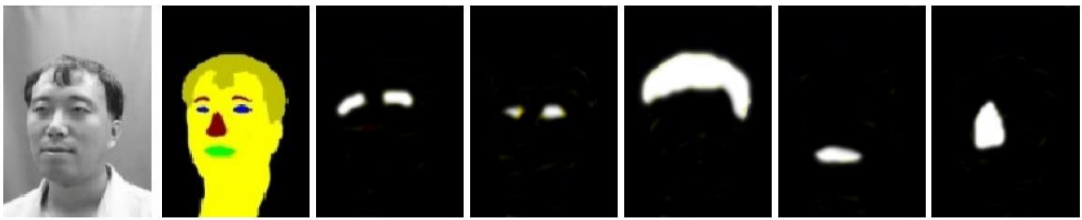

**Figure 1:** Example of segmentation results. Labelled ground truth on the second line, algorithm output on the third.

### *3.3 Race Classification*

As discussed previously, we build a face-parsing model. The face segmentation model produces the most likely class label for every pixel in an image. We created PMs in the segmentation phase for each dense class. Fig. 1 shows a single image segmented using the proposed algorithm. PMs for the five classes that we use for race classification are also shown in Fig. 1. PMs are actually gray scale images in which low intensity at a specific location shows a comparatively lower estimation probability of a specific face class. We use these PMs for race classification in the later stages. We conclude that five of the seven classes help in race classification. Therefore, we use PMs of only these five classes. We represent these PMs as $PMs_{nose}$, $PMs_{back}$, $PMs_{eyes}$, $PMs_{brows}$, $PMs_{skin}$, $PMs_{mouth}$, and $PMs_{hair}$. Figs. 2 and 3 show the original face images from the CAS-PEAL [29] and FERET [25] datasets segmented with the proposed segmentation part. We use information extracted from these segmented images for race classification, whereby we use information from five classes for our race classification part.

**Algorithm 1: Proposed Race Classification Algorithm**

**Input:** $F_{training} = \{(M_n, T_n)\}_{n=1}^{m} = 1$, $F_{testing}$.
where $F_{training}$ is the training data for the deep-learning based model A and $F_{testing}$ is the testing data. $M$ represents
the training face image, where $T(i,j) \in \{0,1,2,3,4,5,6\}$.
**Face Segmentation:**
**a (1):** Training a deep learning-based face parsing model using training data.
**a (2):** Producing segmented faces and *PMs* for each face class
**a (3):** Using the deep learning face model to create *PMs* for all face classes listed as
$PMs_{skin}$, $PMs_{mouth}$, $PMs_{eyes}$, $PMs_{nose}$, $PMs_{brows}$
$PMs_{hair}$, and $PMs_{back}$
**b. Race classification part:**
**b (1):** Information extraction from PMs using DCNNs (only five classes)
**b (2):** Training a classifier A through a feature vector, such that;
$f = PMs_{eyes} + PMs_{mouth} + PMs_{eyebrows} + PMs_{nose} + PMs_{hair}$
where *f* is a unique feature vector for each face.
**Output:** estimated race.

For race classification, we used the extracted facial features from the PMs through DCNNs. After feature extraction, we trained a second classifier for every database. We provided labels for 250 images from each database categorized into seven classes and used these labeled images to build a face segmentation model. We generated PMs for each testing image using the corresponding model.

After performing a large pool of experiments, we investigated the best facial features for improving race classification. We used PMs of only five classes, consisting of hair, eyes, nose, mouth, and brows. We extracted features from these PMs and then combined these features to build a single feature vector. We trained the second stage of a classifier using this PM information. We used 10-fold experiments to validate our model precisely.

During our experiments, we noticed some interesting points. For example, we noticed that minor classes contribute more to race classification than major classes. Hence, we used four minor and one major class for race classification. It is clear from Fig. 1 that PMs for smaller classes differ significantly according to race class. We encoded the extracted information in a vector and used it for race classification. Hair is a major class that has a complex shape geometry; it varies from subject to subject. Previously proposed segmentation methods were not able to extract hair geometry properly. However, our face segmentation part extracts hair information efficiently, as shown in Figs. 1, 2, and 3. The border of hairs in the face image is also detected efficiently.

Skin color and texture information vary from race to race. This variation may be due to the influence of certain environments or genes. We encoded this information in our race classification part and used it as a feature. However, this information does not improve the accuracy of the race classification part. To show the importance of each facial feature in race classification, a feature importance graph is shown in 4. This graph is based on a calculation suggested by [41].

## 4 Results and Discussion

### 4.1 Datasets

We evaluate our race classification part using four datasets, namely CAS-PEAL [29], FERET [25], VNFaces [42], and VMER [43]. Details regarding the datasets are presented in the following paragraphs.

#### 4.1.1 CAS-PEAL

This database has been widely used for different face analysis tasks in the CV literature. We use the database in our current work only for race classification. CAS-PEAL is a large dataset with 99, 594 face images. The number of subjects in CAS-PEAL was also high (1040). Although it is a relatively large database, it has a lower complexity level, making it a comparatively simple database for experiments. Some face images from the CAS-PEAL database are shown in Fig. 5.

#### 4.1.2 FERT

The FERET dataset is used for gender recognition, face recognition, age classification, and race classification. We use the database for race classification. The dataset is collected in a constrained condition, whereby information about each subject is provided. The database is of a medium size, as the number of images is either 14 or 126 only. The number of candidates participating in data collection was 1199. The dataset is available in various versions. We use the color version of the database. Unlike CAS-PEAL, this database is more challenging because variations are present in lighting conditions as well as in facial expressions. Some sample images from the FERET dataset are shown in 6.

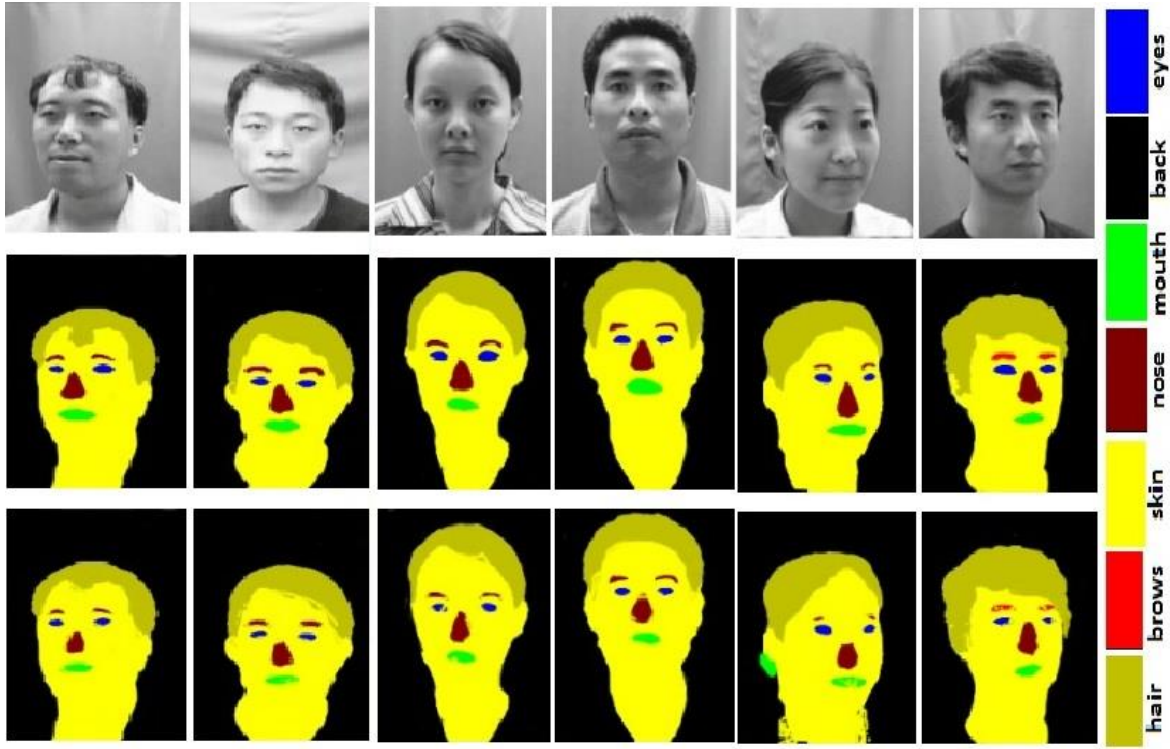

**Figure 2:** CAS-PEAL [29] images: row 1 has original images, row 2 the ground truth, and row 3 is segmented with the proposed segmentation framework.

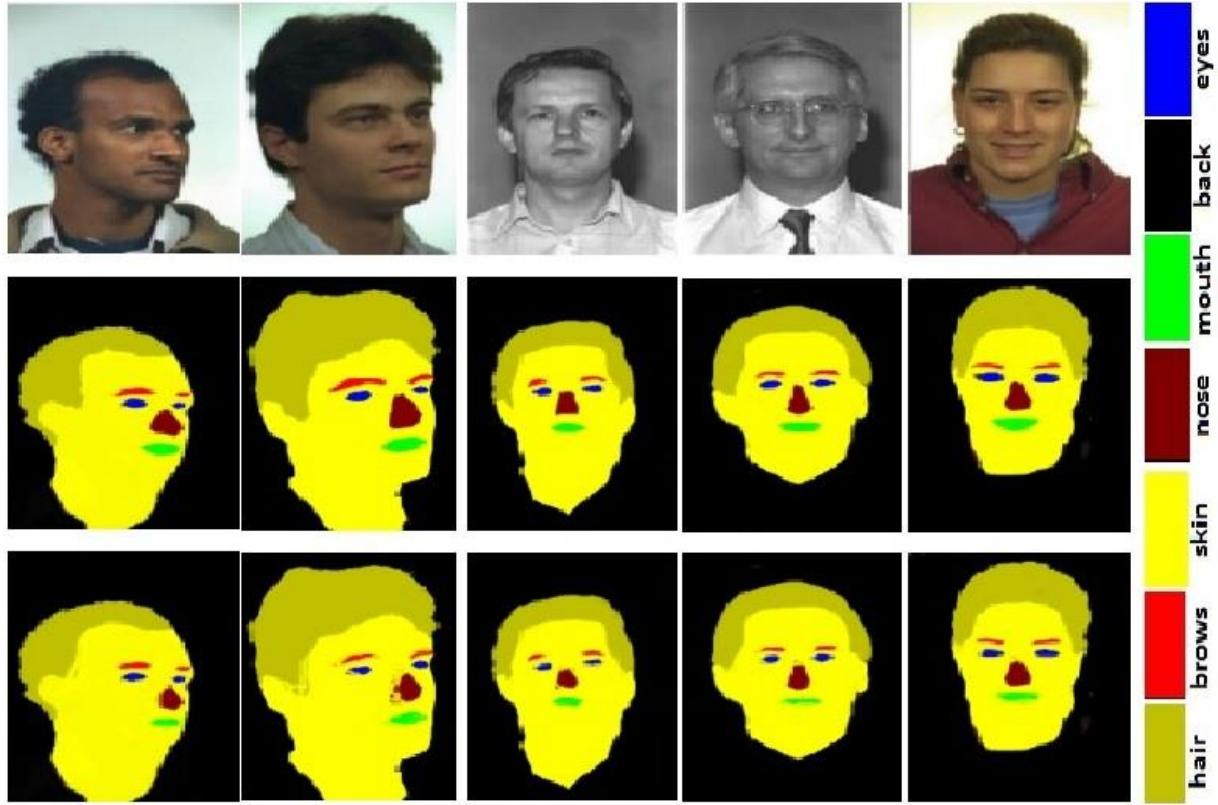

**Figure 3:** Face images from FERET [25] are shown in row 1, labeled images in row 2, and segmented images in row 3.

*4.1.3 VNFaces*

All images in this dataset were collected from the Internet. There is a considerable variation in the age of the candidates in the photos. The images additionally include variations in poses, accessories, illumination conditions, and background scenarios. The Haar cascade is used for automatically detecting the faces. The total number of images collected is 6100, with 2892 Vietnamese faces and 3208 in the category "other". Some example face images from VNFaces are shown in Fig. 7.

*4.1.4 VMER*

All images in this dataset were collected from the Internet. There is a considerable variation in the age of the candidates in the photos. The VMER is the latest database, introduced in 2020. The VMER dataset consists of face images collected from another dataset called VGGFace2 [44]. VGGFace2 is a more comprehensive database with more than 3.3 million images. These images were acquired under diverse lighting conditions. Different occlusion conditions were also introduced in the database. The VMER dataset is divided into four categories: African American, Caucasian Latin, East Asian, and Asian Indian. African Americans have different ethnicity groups, including South American origins, African, and North American. Similarly, East Asia consists of people from China and other South and East Asian regions. Caucasian Latin consists of people from Europe, Western Asia, South America, and North Africa. Lastly, the Asian Indian category contains images of people from India, South Asia, and the Pacific Island areas. A single image showing all these races is shown in Fig. 8.

*4.2 Experiments*

We used an Intel Core i7 CPU during our experiments. The RAM of the system was 16 GB with an NVIDIA 840M GPU. The experimental tools we employed were Keras and TensorFlow. We designed our

DCNN model for 50 epochs with a batch size of 250. The experimental settings for all trained models were the same.

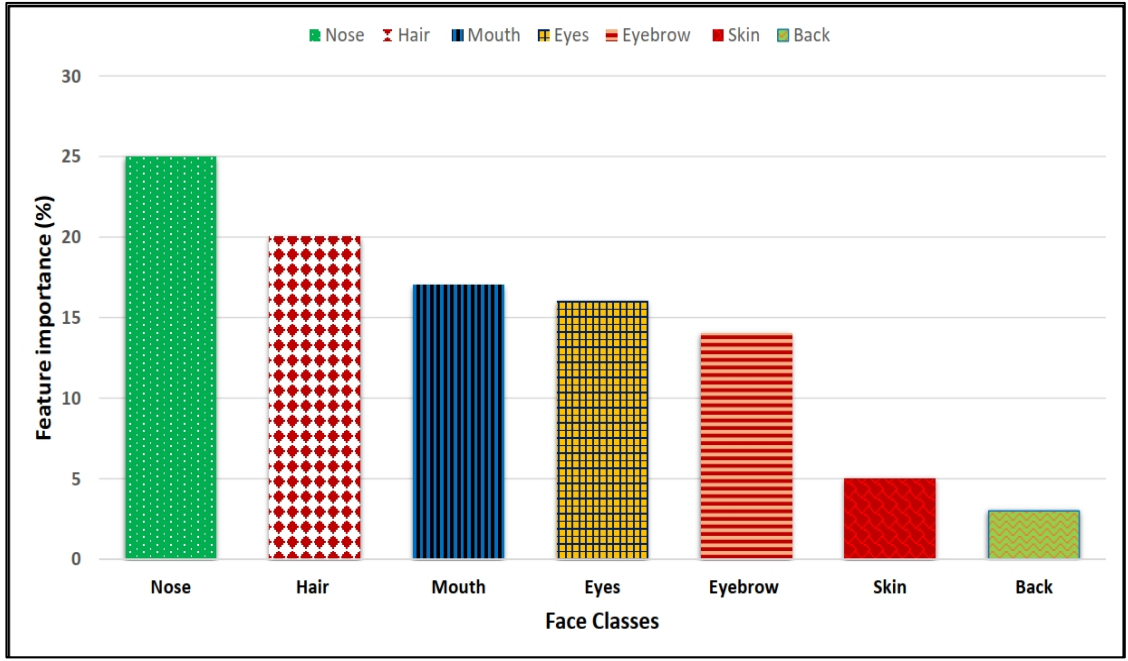

**Figure 4:** Feature importance of face classes calculated through [41].

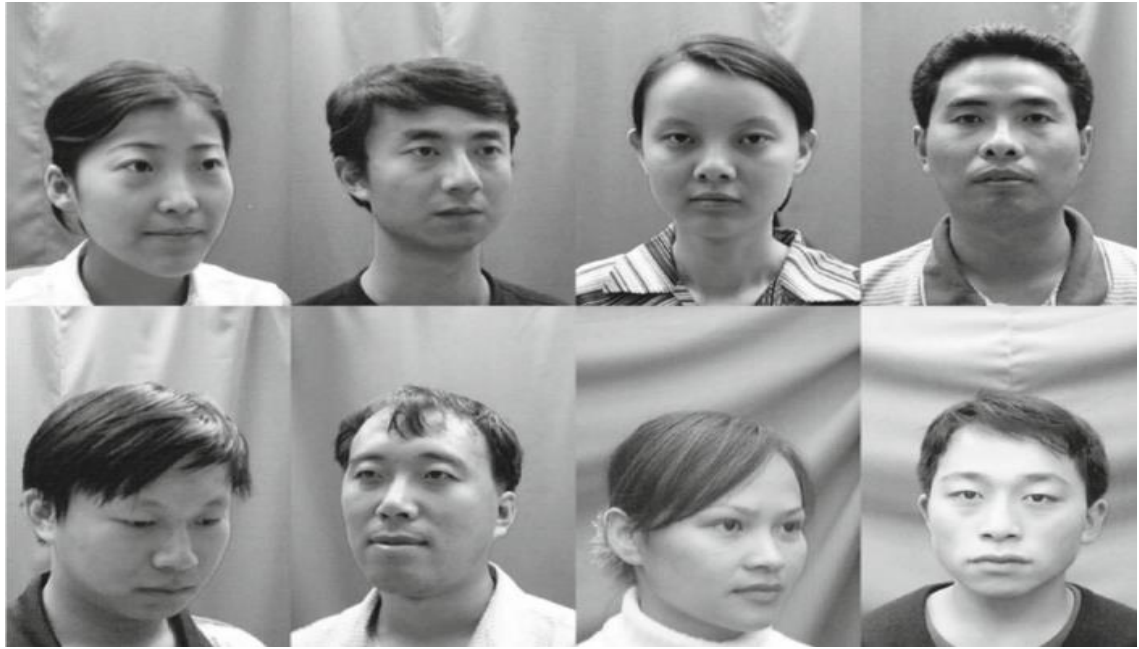

**Figure 5:** Examples of CAS-PEAL face images.

One main drawback of our work is the manual labeling process. It is a time-consuming task and the chances of introducing errors in the labeling process are high. For labeling face images, we use image editing software. Our labeling process depends purely on a single subject involved in the labeling process. If the size of the database to be labeled is large, labeling will be more difficult. We also noticed that the quality of the face images greatly affects the face segmentation part. If the resolution of the images is low, the segmentation results are comparatively poor. Consequently, the performance of the race classification part also decreases.

We observed that minor classes (mouth, eyes, brows, and nose) have a higher contribution to race classification compared with major classes. The PMs produced in the segmentation phase are shown in Fig. 1. We use these PMs as feature descriptors and combine five classes of information for race classification.

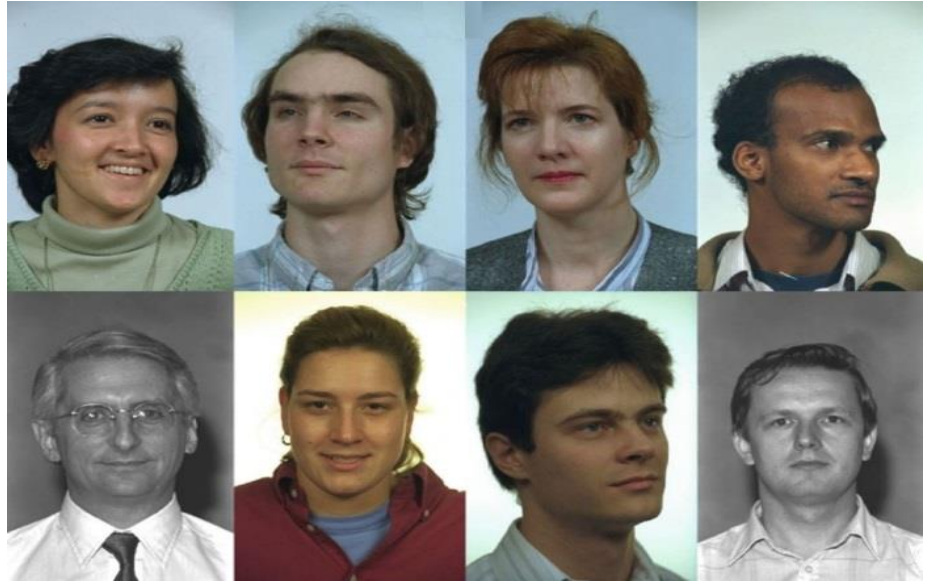

**Figure 6:** Example of FERET [25] face images.

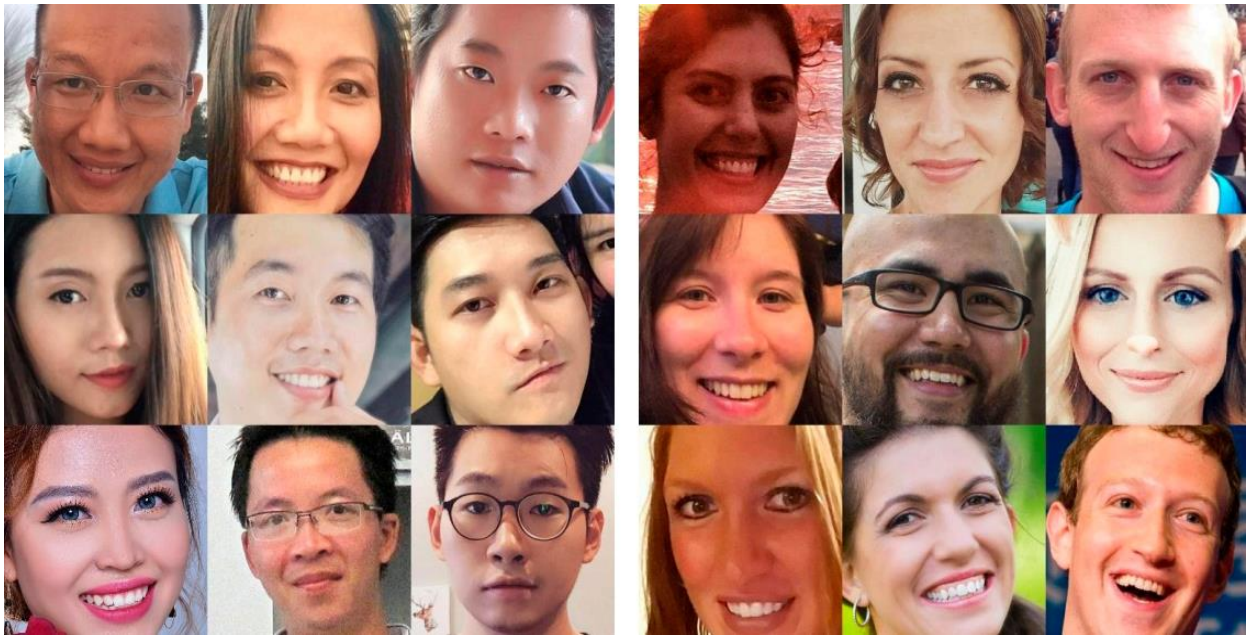

**Figure 7:** Some face images from the VNFaces [43] dataset.

It is important to investigate which facial features help in race classification. It is equally important to know if the sparsity of the features extracted is helpful for feature representation in terms of race classification. We investigate whether racial feature differentiation is particularly conveyed by minor face classes. Hence, we use four minor classes and one major class for race classification. We perform feature importance experiments for investigating important face features for race classification by a adapting strategy from [41]. It is an implementation based on a random forest that calculates which features contribute to a specific classification task. The importance of each face feature in race classification is shown in Fig. 4. It is explicit in Fig. 4 that five face classes contribute more to race classification. These classes include hair, eyebrows, nose, mouth, and eyes. Therefore, we use PMs of only these five classes for race classification. We conduct experiments on four standard race classification datasets: FERET, CAS-PEAL, VNFaces, and VMER. All these datasets are publicly available for research purposes and downloading. We process the captured face images during the first stage. To extract the face part, we use a DCNN-based face detection algorithm proposed in [39]. We also normalize the illumination conditions, as in some face images, the lighting condition is non-uniform.

We manually label 250 face images into seven different classes. This manual labeling process is a time-consuming task. There are also possibilities of error as it is very challenging to differentiate different face regions from each other, for example, differentiating between nose and the skin class. We use 250 manually labeled images for training a DCNN-based face parsing model. We include images from all classes in 250 facial images in each dataset. During our testing phase, we did not include 250 images that were previously used to build the trained model. We create PMs for all seven classes but use only five classes for race classification. Adding all seven classes not only increases the processing time of the algorithm but also reduces the overall race classification accuracy of the model. The results obtained with our model for race classification and its comparison with SOA are shown in Table 3. It is clear from this table that we have achieved better results than previously reported studies for three datasets, excluding VMER. One possible reason for the comparatively poor results for VMER is the resolution and quality of the face images. These images were collected in complex background scenarios, and the lighting conditions are also non-uniform. For these reasons, we have comparatively poor segmentation results for this dataset, leading to a poor performance of the race classification part as well. We investigated a combination of all face features and then used information extracted from only five classes. The validation protocols shown in Table 3 are not the same for all methods. Some of these methods may use different protocols, for instance in the case of [45], a 5-fold cross validation is used. In our experiments, we used a ten-fold validation protocol that validates models more precisely. We obtained classification accuracies of 100%, 99.2%, 92.0%, and 93.2% for the FERET, CAS-PEAL, VNFaces, and VMER datasets, respectively.

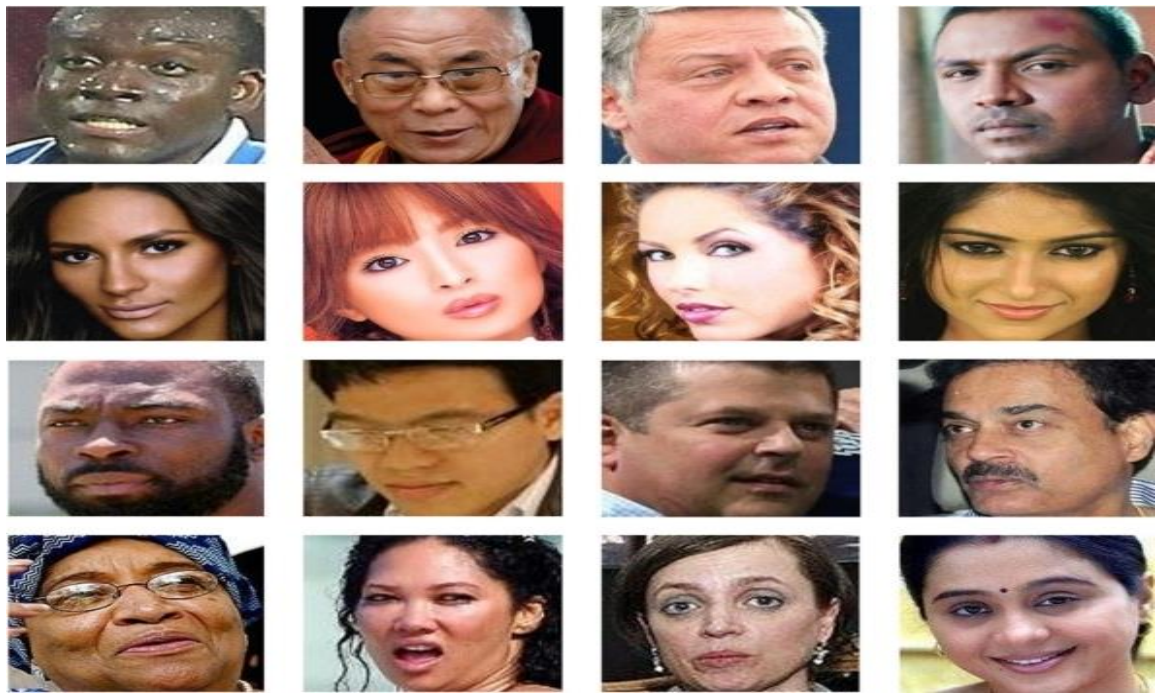

**Figure 8:** Example of VMER [42] face images. First column shows images from African American,

second column East Asian, third column Caucasian Latin, and fourth column Asian Indian images.

**Table 3:** Comparison between proposed method and related works on race classification.

| Authors | Dataset | Method | Race groups | Accuracy (%) |
|---|---|---|---|---|
| Ou et al. [27] | FERET | PCA and SVM | Asian and Non-Asian | 82.5 |
| Muhammed et al. [45] | FERET | LBP and KNN | Eastern/Middle Asian, Hispanic Caucasian, African American | 96.0 |
| Roomi et al. [32] | FERET | skin information and Adaboost | Caucasian, Asian, African, American | 91.6 |
| Masood et al. [46] | FERET | DCNN | Caucasian, Mongolian, and Negroid | 98.6 |
| **Proposed method** | **FERET** | **Facial feature discovery framework** | **Asian and Non-Asian** | **100** |
| Manesh et al. [31] | FERET and CAS-PEAL | Gabor and SVM | Asian and Non-Asian | 98.0 |
| Chen and Ross [47] | CAS-PEAL | Gradient and Gabor pattern | Asian, Caucasian, and African | 98.2 |
| Saliha et al. [28] | CAS-PEAL+FERT | Regression on spark method | Asian and Non-Asian | 99.9 |
| **Proposed method** | **CAS-PEAL** | **Facial feature discovery framework** | **Asian, Caucasian, and African** | **99.2** |
| Vo et al. [42] | VNFaces | DCNNs | Vietnamese and others | 88.87 |
| **Proposed method** | **VNFaces** | **Facial feature discovery framework** | **Vietnamese and others** | **92.0** |
| Greco et al. [43] | VMER | DCNNs | African American, Asian Indian, Caucasian Latin, and East Asian | 94.4 |
| **Proposed method** | **VMER** | **Facial feature discovery framework** | **African American, Caucasian Latin, East Asian, and Asian Indian** | **93.2** |

## 5 Conclusion

In this paper, we introduced a race classification method that utilizes information extracted from a previously introduced face segmentation model. We built a face segmentation framework using DCNNs by extracting information from five different face classes. The face segmentation model provides a semantic class label for every pixel. We used a probabilistic classification method and created PMs for all face classes. We performed experiments to trace useful features for race classification and conclude by using 5 classes. We tested our race classification part on four datasets, including FERET, CAS-PEAL, VNFaces, and VMER. In the future, we plan to optimize the face segmentation part. A crucial point for improving the performance of the face segmentation part is the application of well-managed engineering techniques. One possible combination is exploring CRFs along with DCNNs. Data augmentation and the application of foveated architectures [48] are alternate scenarios to be addressed. Lastly, we provide one route toward other complex face analysis tasks, such as facial gestures and expression recognition, face beautification, and many more. In the future, we will explore these routes as well.

**Funding Statement:** This research was supported by the MSIT (Ministry of Science and ICT), Korea, under the ITRC (Information Technology Research Center) support program (IITP-2020-2018-0-01431) supervised by the IITP (Institute for Information & Communications Technology Planning & Evaluation).

**Conflicts of Interest:** The authors declare that they have no conflicts of interest to report regarding the present study.